\documentclass[runningheads]{llncs}
\usepackage[T1]{fontenc}
\usepackage{graphicx}
\usepackage{booktabs}
\usepackage[misc]{ifsym}

\usepackage{tikz}%

\usetikzlibrary{positioning, fit, backgrounds, arrows.meta}
\usepackage{tabularx}
\usepackage{multirow}%
\newcolumntype{L}[1]{>{\hsize=#1\hsize\raggedright\arraybackslash}X}
\newcolumntype{C}[1]{>{\hsize=#1\hsize\centering\arraybackslash}X}
\newcolumntype{R}[1]{>{\hsize=#1\hsize\raggedleft\arraybackslash}X}

\usepackage{amsmath,amssymb,amsfonts}%
\usepackage{mathrsfs}%
\usepackage[title]{appendix}%
\usepackage{xcolor}%
\usepackage{textcomp}%
\usepackage{manyfoot}%
\usepackage{listings}%
\usepackage{todonotes}%
\usepackage{hyperref}
\usepackage[depth=2]{bookmark}
\usepackage{orcidlink}

\usepackage{pifont}
\newcommand{\cmark}{\ding{51}}%
\newcommand{\xmark}{\ding{55}}%

\usepackage{color-edits}
\addauthor{gemini}{red}
\addauthor{jakub}{blue}
\addauthor{gustav}{magenta}

\usepackage{soul}
\newcounter{mycomment}

\begin{document}

\title{Universal Encoders for Modular\\Relational Deep Learning}

\titlerunning{Universal Encoders for Modular Relational Deep Learning}

\toctitle{Universal Encoders for Modular Relational Deep Learning}

\author{Jakub Pele\v{s}ka\inst{1}~\Letter~\orcidlink{0009-0000-8561-8106} \and
Gustav \v{S}\'{i}r\inst{1}~\orcidlink{0000-0001-6964-4232}}

\authorrunning{J. Peleška and G. Šír}

\institute{Czech Technical University in Prague,\\Karlovo náměstí 13, Prague, 121 35, Czechia\\\email{\{jakub.peleska,gustav.sir\}@cvut.cz}
}

\tocauthor{Jakub Peleška, Gustav Šír}

\maketitle              

\begin{abstract}
Relational Deep Learning (RDL) models multi-tabular data\-bases as temporal heterogeneous graphs for end-to-end representation learning. While RDL is evolving rapidly, existing approaches face signi\-ficant generalization obstacles. They are either schema-specific, requiring training from scratch for every new database, or they rely on monolithic architectures that entangle feature encoding with graph message-passing. Analyzing these limitations, we establish four core \textit{pillars} for building \textit{foundational} relational models: semantic granularity, structural topology, temporal causality, and unified optimization.

Addressing these pillars, we propose a \textit{modular} approach that decouples row encoding from graph message-passing.  For that purpose, we introduce the \textit{Universal Row Encoder} - a transformer-based module that integrates raw cell data with schema metadata---including column semantics, table names, and global distribution statistics---to produce table-width invariant row embeddings. By explicitly feeding global statistics to an intra-row self-attention mechanism, the encoder natively contextualizes unseen features and handles sparse data. Serving as a flexible ``backend'' for any downstream graph architecture, our pretrained encoder enhances cross-database knowledge transfer on the established RelBench benchmarks while improving learning convergence and memory footprint. 

\keywords{Relational Deep Learning  \and Foundational Models \and Relational Databases \and Pretraining}
\end{abstract}

\section{Introduction}
Traditional machine learning pipelines for relational databases (DBs) typically require extensive, manual feature engineering to flatten interconnected tables into a single, static data matrix. In recent years, Relational Deep Learning (RDL)~\cite{cvitkovic_supervised_2020,zahradnik_deep_2023,fey_position_2024} has emerged as a powerful alternative, modeling multi-tabular databases natively as temporal heterogeneous graphs. By representing rows as nodes and primary-foreign key linkages as edges, RDL frameworks operate directly on the raw relational structure by leveraging Heterogeneous Graph Neural Network (HGNN) architectures~\cite{zhang_heterogeneous_2019}. This shift allows for end-to-end representation learning that inherently respects the complex relational topology, underlying the DBs.

However, despite its rapid evolution, most of RDL is still constrained by a severe generalization bottleneck. Standard HGNNs, utilized for RDL, are considerably schema-specific because the input layers and message-passing mechanisms of these models are hardcoded to the specific tables and columns of the training DB; hence, they cannot easily adapt to new databases or new domains. A model trained on a retail database, for instance, cannot be directly transferred to a healthcare database due to disjoint feature spaces and relational structures. Consequently, practitioners are forced to initialize and train a new model from scratch for every new database, limiting the scalability and adoption of RDL.

To overcome these limitations, the RDL community needs to transition towards relational \textit{foundation} models. The overarching goal is to leverage supervised pretraining across a massive, diverse corpus of multi-domain databases to learn universal representations of tabular data and structural motifs. By capturing these broad patterns, a foundational RDL model promises the ability to achieve zero-shot or few-shot transfer on entirely unseen schemas, dramatically reducing both the computational overhead and the volume of labeled data required for new predictive tasks. While there have been some very recent proposals in this direction~\cite{wehrstein_towards_2025,wang_griffin_2025,ranjan_relational_2025}, these current works largely mirror other AI domains by advocating monolithic Transformer-based architectures that entangle feature encoding with graph message passing, blurring the boundary between attribute semantics and relational structure, weakening the key inductive bias salient to the relational DB domain and robust schema transfer.

In this work, we present a two-fold contribution to address these architectural constraints. First, we establish a conceptual framework formalized as the \textit{Four Pillars} of foundational relational modeling: semantic granularity, structural topologies, temporal causality, and unified optimization. These pillars dictate the necessity of a \textit{modular} pretraining framework that explicitly decouples feature encoding from relational message-passing. Second, as our core technical contribution focusing directly on the first pillar, we introduce the \textit{Universal Row Encoder}. This transformer-based module is designed to produce rich, table-width invariant embeddings. By integrating raw cell data with crucial schema metadata---such as column semantics, global distribution statistics, and table names---the encoder natively contextualizes unseen features and robustly handles missing data without flattening the entire database into a global context bottleneck. Crucially, this modular design provides significant architectural flexibility while retaining the original relational learning bias. The pretrained Universal Row Encoder acts as a robust, plug-and-play module for \textit{any} downstream graph architecture (e.g., Heterogeneous Graph Transformer~\cite{hu_heterogeneous_2020} or GraphSAGE~\cite{hamilton_inductive_2017}) and pairs seamlessly with diverse decoders, ranging from task-specific multi-layer perceptrons to unified In-Context Learning (ICL)~\cite{brown_language_2020} modules. The Universal Row Encoder has been implemented as a component of the \textsc{ReDeLEx} framework~\cite{peleska_redelex_2025}, and its source code is readily available on GitHub\footnote{\url{https://github.com/jakubpeleska/redelex}}.

\section{The Pillars of the Relational Foundation Models}

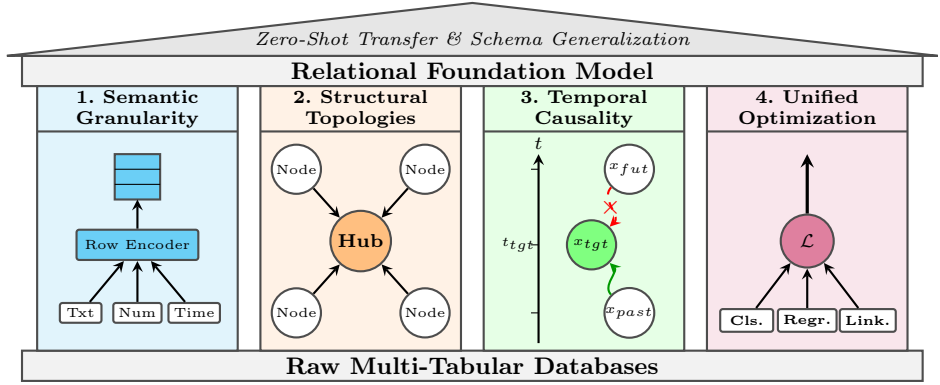
\begin{figure}[t]
    \centering
    \begin{tikzpicture}[
    >=stealth,
    pillar/.style={draw=black!70, thick, fill=white},
    pillarhead/.style={text width=2.4cm, align=center, font=\scriptsize\bfseries, minimum height=0.7cm},
    component/.style={draw=black!70, thick, fill=white, align=center, font=\tiny, rounded corners=1pt},
    c1/.style={fill=cyan!50},
    c2/.style={fill=orange!50},
    c3/.style={fill=green!50},
    c4/.style={fill=purple!50}
]

    \draw[thick, draw=black!70, fill=gray!10] (0.0, 0.0) rectangle (11.9, 0.4);
    
    \node at (5.95, 0.2) {\textbf{Raw Multi-Tabular Databases}};

    \draw[thick, draw=black!70, fill=gray!10] (0.0, 3.9) rectangle (11.9, 4.3);
    \node at (5.95, 4.1) {\textbf{Relational Foundation Model}};
    \draw[thick, draw=black!70, fill=gray!20] (-0.2, 4.3) -- (5.95, 5.0) -- (12.1, 4.3) -- cycle;
    \node[font=\scriptsize\itshape] at (5.95, 4.5) {Zero-Shot Transfer \& Schema Generalization};

    \draw[pillar, fill=cyan!10] (0.2, 0.4) rectangle (2.85, 3.9);
    \draw[thick, draw=black!70] (0.2, 3.3) -- (2.85, 3.3);
    \node[pillarhead] at (1.525, 3.6) {1. Semantic\\Granularity};
    
    \node[component, inner sep=2pt, minimum width=0.55cm] (t1) at (0.775, 0.9) {Txt};
    \node[component, inner sep=2pt, minimum width=0.55cm] (t2) at (1.525, 0.9) {Num};
    \node[component, inner sep=2pt, minimum width=0.55cm] (t3) at (2.275, 0.9) {Time};
    
    \node[component, c1, minimum width=1.6cm, minimum height=0.4cm] (enc) at (1.525, 1.8) {Row Encoder};
    
    \draw[thick, draw=black!70, fill=cyan!50] (1.225, 2.4) rectangle (1.825, 3.0);
    \draw (1.225, 2.6) -- (1.825, 2.6);
    \draw (1.225, 2.8) -- (1.825, 2.8);
    
    \draw[->, thick] (t1) -- (enc);
    \draw[->, thick] (t2) -- (enc);
    \draw[->, thick] (t3) -- (enc);
    \draw[->, thick] (enc.north) -- (1.525, 2.4);

    \draw[pillar, fill=orange!10] (3.15, 0.4) rectangle (5.8, 3.9);
    \draw[thick, draw=black!70] (3.15, 3.3) -- (5.8, 3.3);
    \node[pillarhead] at (4.475, 3.6) {2. Structural\\Topologies};
    
    \node[component, c2, circle, minimum size=0.8cm, inner sep=0pt, font=\scriptsize\bfseries] (hub) at (4.475, 1.85) {Hub};
    \node[component, circle, minimum size=0.65cm, inner sep=0pt] (s1) at (3.625, 0.9) {Node};
    \node[component, circle, minimum size=0.65cm, inner sep=0pt] (s2) at (5.325, 0.9) {Node};
    \node[component, circle, minimum size=0.65cm, inner sep=0pt] (s3) at (3.625, 2.8) {Node};
    \node[component, circle, minimum size=0.65cm, inner sep=0pt] (s4) at (5.325, 2.8) {Node};
    
    \draw[->, thick] (s1) -- (hub);
    \draw[->, thick] (s2) -- (hub);
    \draw[<-, thick] (hub) -- (s3);
    \draw[<-, thick] (hub) -- (s4);

    \draw[pillar, fill=green!10] (6.1, 0.4) rectangle (8.75, 3.9);
    \draw[thick, draw=black!70] (6.1, 3.3) -- (8.75, 3.3);
    \node[pillarhead] at (7.425, 3.6) {3. Temporal\\Causality};
    
    \draw[->, thick] (6.825, 0.5) -- (6.825, 3.0);
    \node[font=\scriptsize\bfseries] at (6.825, 3.15) {$t$};
    \draw (6.725, 0.9) -- (6.825, 0.9);
    \draw (6.725, 1.8) -- (6.825, 1.8) node[left, font=\tiny, inner sep=1pt] {$t_{tgt}$};
    \draw (6.725, 2.8) -- (6.825, 2.8);

    \node[component, circle, minimum size=0.65cm, inner sep=0pt] (p_past) at (8.025, 0.9) {$x_{past}$};
    \node[component, circle, minimum size=0.65cm, inner sep=0pt, c3, thick] (p_tgt) at (7.525, 1.8) {$x_{tgt}$};
    \node[component, circle, minimum size=0.65cm, inner sep=0pt] (p_fut) at (8.025, 2.8) {$x_{fut}$};
    
    \draw[->, green!60!black, thick] (p_past) to[out=135, in=-45] (p_tgt);
    \draw[->, red, thick, dashed] (p_fut) to[out=-135, in=45] (p_tgt);
    
    \node[red, font=\normalsize\bfseries] at (7.775, 2.3) {$\times$};

    \draw[pillar, fill=purple!10] (9.05, 0.4) rectangle (11.7, 3.9);
    \draw[thick, draw=black!70] (9.05, 3.3) -- (11.7, 3.3);
    \node[pillarhead] at (10.375, 3.6) {4. Unified\\Optimization};
    
    \node[component, inner sep=2pt, minimum width=0.7cm, minimum height=0.3cm,font=\tiny\bfseries] (task1) at (9.575, 0.8) {Cls.};
    \node[component, inner sep=2pt, minimum width=0.7cm, minimum height=0.3cm,font=\tiny\bfseries] (task2) at (10.375, 0.8) {Regr.};
    \node[component, inner sep=2pt, minimum width=0.7cm, minimum height=0.3cm, font=\tiny\bfseries] (task3) at (11.175, 0.8) {Link.};
    
    \node[component, c4, circle, minimum size=0.7cm, inner sep=0pt, font=\scriptsize\bfseries] (loss) at (10.375, 1.85) {$\mathcal{L}$};
    
    \draw[->, thick] (task1) -- (loss);
    \draw[->, thick] (task2) -- (loss);
    \draw[->, thick] (task3) -- (loss);
    
    \draw[->, thick, line width=1.2pt] (loss.north) -- (10.375, 3.0);

\end{tikzpicture}
    \caption{The Four Pillars of Relational Foundation Models. A modular architecture is required to systematically resolve semantic granularity, structural topologies, temporal causality, and unified optimization without entangling the representational space.}
    \label{fig:pillars}
\end{figure}

To successfully pretrain a foundational RDL model that generalizes across diverse databases, we propose mapping the fundamental properties of relational data. We establish four core ``pillars'' of relational modeling (visualized in Figure~\ref{fig:pillars}). These pillars introduce complexities that cannot be directly resolved by monolithic architectures or existing flattening (propositionalization) techniques~\cite{kramer2001propositionalization}. Attempting to force these challenges into a single, entangled learning representation inevitably leads to memory bottlenecks, temporal leakage, or loss of transferability. Instead, resolving these pillars motivates a modular approach, in which specialized components handle specific complexities before integrating into a unified architecture.

\subsection{Semantic \& Feature Granularity}
\subsubsection{Cross-Table Feature Distribution}
Databases contain diverse data types (continuous, categorical, text, timestamps) with different scales. A foundational model must map these schema-specific signals into a shared, universal latent space without relying on hardcoded input dimensions~\cite{huang2020tabtransformer}.

\subsubsection{Intra-Table Feature Relationships}
Before an entity can share information across a graph, its own local context must be understood. Intra-row relationships are semantic (e.g., ``price'' and ``discount'' must be contextualized together). Architectures that treat every cell as a distinct node (token) in a massive global graph context can dilute these high-density intra-row relationships.

\subsubsection{Robustness to Sparse Data}
Real-world relational DBs are rarely clean; they are sparse and frequently contain missing values. A foundational model requires objectives that teach it to impute missing values and reason over incomplete relational contexts, which calls for dedicated feature-level processing mechanisms.

\subsection{Structural \& Relational Topologies}
\subsubsection{The Semantics of PK-FK Relationships}
Not all edges in a relational graph are equivalent. A Primary Key-Foreign Key (PK-FK) linkage might represent a ``belongs-to'' relationship or a ``happened-at'' relationship. The model must distinguish between these relational semantics to route information correctly.

\subsubsection{Overcoming Structural Bottlenecks}
Relational schemas often form ``star'' schemas, creating massive hub nodes. Standard message-passing GNNs can suffer from over-smoothing and neighborhood explosion when traversing these hubs~\cite{hamilton_inductive_2017}. Foundational models require structural routing to navigate such bottlenecks without losing signal fidelity.

\subsubsection{Strict Schema-Agnosticism}
The ultimate goal of a foundation model is transferability. The architecture must be agnostic to the graph topology and node types of the pretraining data, enabling it to ingest novel relational structures during inference without requiring structural retraining.

\subsection{Temporal Causality}
\subsubsection{Continuous Time Representation and Strict Causality}
Most real-world DBs are living records of events over time. Treating time as a continuous, causal dimension is critical to prevent \textit{temporal leakage} during message passing. Monolithic entangled spatio-temporal graphs easily violate causality unless temporal constraints are decoupled and enforced.

\subsubsection{Asynchronous Multi-Table Dynamics}
In real-world databases, temporal events occur at vastly different frequencies across different tables. A \texttt{users} table might update rarely, while a \texttt{clicks} or \texttt{transactions} table generates thousands of rows per second. A foundational model must be able to align these asynchronous timelines, avoiding the over-representation of high-frequency tables while retaining crucial low-frequency state changes.

\subsubsection{Temporal Concept Drift and Feature Evolution}
Entities in a database are not static; their behaviors and data distributions shift over time (concept drift)~\cite{gama2014survey}. A foundational model must learn temporally adaptive representations. It requires mechanisms to decay outdated information and weigh recent context appropriately, rather than treating history as a flat set of static edges.

\subsection{Unified Optimization \& Transferability}

\subsubsection{Unified Task Representation}
Supervised pretraining on relational data requires learning from vastly different tasks (e.g., predicting user churn, forecasting sales, classifying item categories), projecting disjoint label spaces into a unified objective. This requires specialized decoders, such as text-encoded multi-task cross-attention modules or appending predictive tasks as ``task tables,''~\cite{fey_position_2024} to unify all training signals into a standard masked token prediction problem.

\subsubsection{Uniform Representation of Sampled Subgraphs}
Because graph sampling is necessary for scale, the model must be robust to the variance introduced by sampling different temporal subgraphs. Standardizing the input spaces of these highly skewed relational graphs is necessary to ensure stable optimization.

\subsubsection{Balancing Loss and Gradient Scales}
Pretraining across diverse supervised tasks (regression vs. classification, high-frequency vs. low-frequency labels) creates conflicting gradient signals. A foundational architecture must modularize its learning objectives to balance gradient contributions and prevent dominant tasks from causing catastrophic interference with the generalized representation.

\begin{figure}[t]
    \centering
    \resizebox{\textwidth}{!}{%
        \begin{tikzpicture}[
        node distance=1.5cm,
        module/.style={draw, rounded corners, align=center, minimum height=1.6cm, minimum width=2.4cm, font=\small\bfseries},
        data/.style={draw, rounded corners, align=center, minimum height=0.8cm, minimum width=2cm, font=\footnotesize},
        arrow/.style={->, >=stealth, thick, rounded corners=2pt}
    ]

    \node[data, fill=gray!10] (db) {Arbitrary Database\\with\\Prediction Task};

    \node[module, fill=green!10, right=0.5cm of db] (encoder) {Universal\\Row Encoder};

    \node[module, fill=purple!10, right=2cm of encoder] (gnn) {Graph Neural\\Architecture};

    \node[module, fill=orange!10, right=2cm of gnn] (head) {Prediction\\Head};

    \node[data, fill=gray!10, right=0.5cm of head] (output) {Output\\Prediction};

    \draw[arrow] (db.east) -- (db.east -| encoder.west);
    
    \draw[arrow] (encoder.east) -- node[above, font=\scriptsize, align=center] {Uniform Node\\Features $\in \mathbb{R}^d$} (gnn.west);
    \draw[arrow] (gnn.east) -- node[above, font=\scriptsize, align=center] {Contextualized\\Embeddings} (head.west);
    \draw[arrow] (head.east) -- (output.west);

    \begin{scope}[on background layer]
        \node[draw, dashed, fill=blue!4, rounded corners, fit=(encoder) (head), inner xsep=0.3cm, inner ysep=0.3cm] (framework_bg) {};
        \node[anchor=south west, font=\itshape\small, text=black!80] at (framework_bg.north west) {Modular RDL Framework};
    \end{scope}

\end{tikzpicture}
    }
    \caption{The modular framework for Relational Deep Learning. The architecture explicitly decouples semantic feature extraction from structural message passing. The Universal Row Encoder standardizes arbitrary database rows into uniform node features, which are then processed by a graph neural network, and finally mapped to predictions via task-specific decoders.}
    \label{fig:modular_framework}
\end{figure}
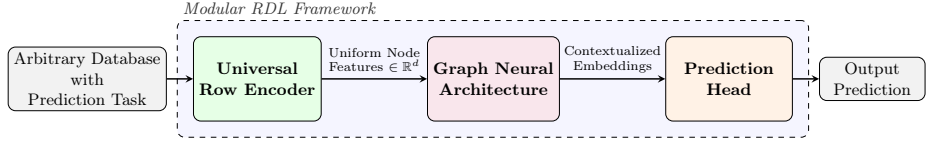

\section{Modular Pretraining for RDL}
To address the four pillars outlined above, we propose moving away from monolithic RDL architectures that attempt to solve feature mapping, temporal routing, and message passing simultaneously. Attempting to pass every cell of a database into a global graph transformer~\cite{yun_graph_2019} results in an intractable $\mathcal{O}(N^2)$ memory bottleneck, while standard Heterogeneous GNNs fail to generalize across schemas.

Instead, we adopt a \textit{divide-and-conquer} formulation. Let a database be a collection of tables $\mathcal{D}=\{T_k\}_{k=1}^{K}$ with schemas $\mathcal{C}_k=\{c^{(k)}_j\}_{j=1}^{m_k}$ and rows $\{r^{(k)}_i\}_{i=1}^{n_k}$. We map $\mathcal{D}$ to a temporal relational graph $G=(V,E,\tau)$ where each row corresponds to a node $v=(k,i)\in V$, PK--FK links induce edges $(v\rightarrow u)\in E$, and $\tau:V\rightarrow\mathbb{R}$ assigns event times.

Our modular pipeline then factorizes the predictor into three components:
\begin{equation}
    h_v = f_{\theta}(r_v) \in \mathbb{R}^d, \qquad \{z_v\}_{v\in V_S} = g_{\phi}\big(G_S,\{h_v\}_{v\in V_S}\big), \qquad \hat{y}_u = d_{\psi}(z_u),
\end{equation}
where $f_{\theta}$ is a universal row encoder (schema-agnostic interface via fixed $d$), $g_{\phi}$ is a graph module applied to a sampled (temporal) subgraph $G_S$, and $d_{\psi}$ is a task head. This strict separation of concerns then decouples semantic feature extraction (Pillar~1) from structural and temporal propagation (Pillars~2--3), while enabling flexible multi-task optimization (Pillar~4).

\subsection{The Universal Row Encoder and Statistical Contextualization}

At the foundation of our modular pipeline is the \textit{Universal Row Encoder}. Its objective is to resolve {semantic and feature granularity} by transforming any multi-typed database row into a fixed-size embedding $h \in \mathbb{R}^d$. Let a row from a table with $m$ columns be represented as
\begin{equation}
    r = \big\{(v_j, c_j, t, s_j, \delta_j)\big\}_{j=1}^{m},
\end{equation}
where $v_j$ is the raw cell value, $c_j$ is the column identifier (e.g., name), $t$ is the table identifier (e.g., name), $s_j$ are precomputed column statistics, and $\delta_j\in\{0,1\}$ indicates missingness. 
Specifically, the contents of the precomputed column statistics $s_j$ depend on the semantic type of the feature $j$: numerical columns utilize the mean, standard deviation, minimum, maximum, median, and the first and third quartiles; categorical and multi-categorical columns are summarized by their overall cardinality alongside precomputed text embeddings of their unique values; and timestamp columns capture temporal boundaries by explicitly storing the earliest and latest exact dates as well as the minimum and maximum years. For temporal databases, all statistics $s_j$ are computed in a split-aware manner (i.e., on the training portion, or only from past data for causal evaluation) to avoid leaking information from the future into the encoder.

\begin{figure}[t]
    \centering
    \resizebox{\textwidth}{!}{%
        \begin{tikzpicture}[
        node distance=0.3cm and 0.3cm,
        box/.style={draw, rounded corners, fill=blue!5, align=center, minimum height=0.8cm, minimum width=2cm, font=\scriptsize},
        enc/.style={draw, rounded corners, fill=green!5, align=center, minimum height=0.8cm, minimum width=2cm, font=\scriptsize},
        pool/.style={draw, rounded corners, fill=orange!5, align=center, minimum height=0.8cm, minimum width=1.5cm, font=\scriptsize},
        cell/.style={draw, rounded corners, fill=purple!10, align=center, minimum height=0.8cm, minimum width=1.5cm, font=\scriptsize},
        longbox/.style={draw, rounded corners, align=center, minimum height=4.5cm, minimum width=1.0cm, inner sep=0.1cm, font=\scriptsize},
        sum/.style={draw, circle, inner sep=0pt, minimum size=0.4cm, fill=yellow!10, font=\scriptsize},
        arrow/.style={->, >=stealth, thick, rounded corners=2pt}
    ]

    \node[box] (raw) {Raw Cell};
    \node[box, below=of raw] (stats) {Column\\Stats};
    \node[box, below=of stats] (name) {Schema\\Metadata};
    \node[box, below=of name] (stype) {Semantic\\Type};

    \node[enc, right=of raw] (val_enc) {Type-Specific\\Encoder};
    \node[enc, right=of stats] (stats_enc) {Stats Linear\\Projection};
    \node[enc, right=of name] (name_enc) {Text\\Embedder};
    \node[enc, right=of stype] (stype_enc) {Type\\Embedding};

    \draw[arrow] (raw) -- (val_enc);
    \draw[arrow] (stats) -- (stats_enc);
    \draw[arrow] (name) -- (name_enc);
    \draw[arrow] (stype) -- (stype_enc);

    \path (stats_enc.east) -- (name_enc.east) coordinate[midway] (mid_enc);
    \node[sum, right=0.5cm of mid_enc] (plus) {$\bigoplus$};

    \draw[arrow] (val_enc.east) to[out=0, in=120] (plus.120);
    \draw[arrow] (stats_enc.east) to[out=0, in=160] (plus.160);
    \draw[arrow] (name_enc.east) to[out=0, in=200] (plus.200);
    \draw[arrow] (stype_enc.east) to[out=0, in=240] (plus.240);

    \begin{scope}[on background layer]
        \node[draw, dashed, fill=gray!5, rounded corners, fit=(raw) (stype_enc) (plus), minimum height=4.5cm, inner xsep=0.2cm] (col_bg) {};
        \node[anchor=south west, font=\itshape\scriptsize, text=black!70] at (col_bg.north west) {Per-Column Statistical Contextualization};
    \end{scope}

    \coordinate[right=1.4cm of plus] (cell_center);
    
    \begin{scope}[on background layer]
        \node[draw, dashed, fill=gray!5, rounded corners, minimum height=4.5cm, minimum width=1.8cm] at (cell_center) (row_bg) {};
        \node[anchor=south, font=\itshape\scriptsize, text=black!70] at (row_bg.north) {Row Cells};
    \end{scope}

    \node[cell] at ([yshift=1.4cm]row_bg.center) (h1) {$h_1$};
    \node[cell] at ([yshift=0.4cm]row_bg.center) (h2) {$h_2$};
    \node[font=\scriptsize] at ([yshift=-0.5cm]row_bg.center) (vdots) {$\vdots$};
    \node[cell] at ([yshift=-1.4cm]row_bg.center) (hn) {$h_n$};

    \draw[arrow, dashed] (plus) to[out=0, in=180] (h2.west);
    \draw[arrow, dashed] (plus) to[out=0, in=180] (hn.west);
    \draw[arrow] (plus) to[out=0, in=180] (h1.west);

    \node[longbox, fill=purple!15, right=0.4cm of row_bg] (transformer) {\rotatebox{90}{\begin{tabular}{c}\textbf{Transformer Aggregator} \\ \textcolor{black!70}{\scriptsize(Intra-Row Self-Attention)}\end{tabular}}};

    \draw[arrow] (h1.east) -- (h1.east -| transformer.west);
    \draw[arrow] (h2.east) -- (h2.east -| transformer.west);
    \draw[arrow] (hn.east) -- (hn.east -| transformer.west);

    \path (transformer.east) coordinate (trans_mid);
    \node[pool, right=0.4cm of trans_mid] (sum_pool) {Sum Pool};
    \node[pool, above=0.4cm of sum_pool] (max_pool) {Max Pool};
    \node[pool, below=0.4cm of sum_pool] (attn_pool) {Attn. Pool};

    \draw[arrow] (transformer.east |- max_pool.west) -- (max_pool.west);
    \draw[arrow] (transformer.east |- sum_pool.west) -- (sum_pool.west);
    \draw[arrow] (transformer.east |- attn_pool.west) -- (attn_pool.west);

    \node[longbox, fill=blue!5, right=0.4cm of sum_pool] (concat) {\rotatebox{90}{Concat \& Projection}};
    \node[longbox, fill=red!10, right=0.4cm of concat] (output) {\rotatebox{90}{\textbf{Row Embedding $\in \mathbb{R}^d$}}};

    \draw[arrow] (max_pool.east) -- (max_pool.east -| concat.west);
    \draw[arrow] (sum_pool.east) -- (sum_pool.east -| concat.west);
    \draw[arrow] (attn_pool.east) -- (attn_pool.east -| concat.west);

    \draw[arrow] (concat.east) -- (output.west);

\end{tikzpicture}
    }
    \caption{Architecture of the Universal Row Encoder. Raw cell values are embedded and additively contextualized with distribution statistics, schema metadata, and semantic types. Contextualized column embeddings are aggregated via an intra-row transformer and a pooling mechanism to produce a fixed-size row representation.}
    \label{fig:row_encoder}
\end{figure}
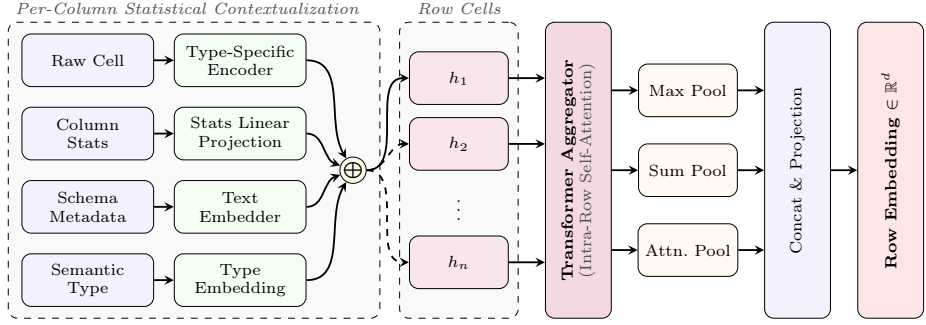

For each column $j$, we construct a contextualized column embedding $x_j \in \mathbb{R}^d$ as a sum of type-specific value embeddings and metadata embeddings:
\begin{equation} \label{eq:contextualization}
    x_j = f_{\mathrm{stype}(j)}(v_j) + g_{\mathrm{name}}(c_j) + g_{\mathrm{table}}(t) + g_{\mathrm{stats}}(s_j) + g_{\mathrm{miss}}(\delta_j).
\end{equation}
Here, $f_{\mathrm{stype}(j)}$ denotes a semantic-type encoder. In our implementation, numerical values are standardized using $s_j$ and embedded by a small MLP; categorical values are mapped to trainable embeddings and combined with an embedding of the raw token; timestamps are embedded via a learned projection of normalized time (e.g., scaled to the $[\min,\max]$ range in $s_j$) and a missingness flag. The functions $g_{\mathrm{name}}$ and $g_{\mathrm{table}}$ create representation of schema text using a text summarization;\footnote{In the reported experiments we used pooled GloVe~\cite{pennington2014glove} embedding vectors, though more powerful models could be employed to further improve performance.} $g_{\mathrm{stats}}(s_j)$ and $g_{\mathrm{miss}}(\delta_j)$ are lightweight learned projections that inject distributional context and explicit missingness.

The sequence $(x_1,\dots,x_m)$ is then processed by an intra-row transformer to obtain contextualized token representations $(z_1,\dots,z_m)$:
\begin{equation}
    (z_1,\dots,z_m) = \mathrm{Transformer}(x_1,\dots,x_m).
\end{equation}

Here, we use \textit{no positional encodings} for columns. As a result, the encoder treats the row as a set of column tokens, making the transformation permutation-equivariant with respect to column order\footnote{While this design ensures transferability across diverse databases by preventing the model from using a specific table layout as a learning \textit{shortcut}, it inevitably discards any latent information encoded in the original column order, such as implicitly ordered feature sequences or deliberately denormalized schemas.} and avoiding dependence on arbitrary schema ordering~\cite{zaheer2017deepsets,lee2019settransformer}. 

Finally, we derive a fixed-size row embedding by pooling over columns. Concretely, we use sum pooling, max pooling, and attentive pooling,
\begin{equation}
    p_{\mathrm{sum}} = \sum_{j=1}^{m} z_j, \quad p_{\mathrm{max}} = \max_{j=1}^{m} z_j, \quad p_{\mathrm{att}} = \sum_{j=1}^{m} \alpha_j z_j, \; \alpha_j = \mathrm{softmax}(q^\top z_j),
\end{equation}
and project their concatenation to $\mathbb{R}^d$:
\begin{equation}\label{eq:pooling-strategies}
    h = W\,[p_{\mathrm{sum}}\,\|\,p_{\mathrm{max}}\,\|\,p_{\mathrm{att}}].
\end{equation}
This design yields a uniform representation per row (independent of table width at the interface to downstream graph layers), while the intra-row attention cost depends only on the number of columns $m$ in the processed table (not on the number of rows $n$ or the global database size, avoiding the $O(N^2)$ bottleneck).

\subsection{Graph-Agnostic Structural Modeling}

After row encoding, each node $v\in V$ is represented by $h_v\in\mathbb{R}^d$, yielding a unified feature space shared across tables. A downstream graph module $g_{\phi}$ then performs message passing on a sampled subgraph $G_S=(V_S,E_S)$ to produce structural representations $z_v$. 
To prevent temporal leakage, we enforce a causal constraint in sampling and propagation: for any prediction target node $u$ at time $\tau(u)$, all nodes and edges used for its computation satisfy
    $(v\rightarrow u)\in E_S \;\Rightarrow\; \tau(v) \leq \tau(u)$,
and more generally the sampled neighborhood must not include events from the future relative to the target time. This temporal constraint is applied when constructing $G_S$ (e.g., in a temporally filtered neighbor loader) before any GNN layers are evaluated.

Note how the modular interface provides the {architectural flexibility}, where the same row embeddings can be used with homogeneous or heterogeneous GNN backbones. In a heterogeneous model, the message-passing parameters may depend on the edge types; in a homogeneous model, parameters are independent of the original schema.

\subsection{Flexible Decoding}

To address the challenges of {unified optimization} (Pillar 4), the modular framework extends to the decoder. Because the graph module outputs standardized node representations $z_u$, the pipeline naturally supports multi-task learning with task-specific heads. Let $\{\mathcal{T}_\ell\}_{\ell=1}^{L}$ denote a set of supervised tasks, each defining a distribution over labeled nodes $(u,y)\sim\mathcal{T}_\ell$ and a loss function $\mathcal{L}_\ell$. We optimize
\begin{equation}
    \min_{\theta,\phi,\{\psi_\ell\}} \; \sum_{\ell=1}^{L} \lambda_\ell\, \mathbb{E}_{(u,y)\sim\mathcal{T}_\ell}\Big[\mathcal{L}_\ell\big(d_{\psi_\ell}(z_u),y\big)\Big],
\end{equation}
where $\lambda_\ell$ balances gradient contributions across tasks.

\subsection{Supervised Pretraining Strategies}

The outlined modularity then enables two distinct supervised pretraining regimes:

\subsubsection{Encoder-Only Pretraining}
In this regime, \textit{only} the Universal Row Encoder is pretrained across diverse databases. It is then frozen and deployed as a general-purpose neural feature extractor. This reduces the computational overhead of training RDL on new databases from scratch, while replacing manual feature engineering with robust, pretrained embeddings.

\subsubsection{End-to-End Pretraining}
In this regime, the Universal Row Encoder, the GNN architecture, and the task heads are pretrained simultaneously. This allows the row encoder to receive gradients shaped by structural and temporal motifs in the downstream graph module. In principle, such end-to-end pretraining can improve transfer to previously unseen relational structures, compared to pretraining the encoder in isolation.

\section{Experiments}
The experiments are designed to assess the framework's modularity and cross-database transferability enabled by the Universal Row Encoder. We follow the standard predictive RDL setup, where each supervised task $\mathcal{T}_\ell$ is defined by (i) a set of target nodes $V^{(\ell)}_{\mathrm{target}}\subseteq V$ (e.g., rows of a specific table), (ii) a label function $y^{(\ell)}:V^{(\ell)}_{\mathrm{target}}\rightarrow\mathcal{Y}_\ell$, and (iii) a prediction time $t_u=\tau(u)$ (optionally with a fixed horizon) that determines the causal subgraph used for computing $z_u$. For evaluation, we report a task-specific metric $m_\ell(\hat{y}_u,y_u)$ (ROC AUC for classification and MAE for regression in our setting) aggregated over held-out targets. The source code for all experiments can be accessed on GitHub\footnote{\url{https://github.com/jakubpeleska/redelex/tree/main/experiments/universal_encoder}}.

\begin{table}[t]
    \centering
    \begin{tabular}{p{2.5cm}ccccccccccc}
\hline
\toprule
Database & Domain & \#Tables & \#FK & \#Factual  & Diameter & Cycle & 1:1 & 1:N \\ 
\midrule
\multicolumn{9}{c}{CTU Relational Databases} \\ 
\midrule
employee & Retail & 6 & 6 & 16 & 3 & \cmark & 2 & 5 \\
ergastf1 & Sport & 13 & 19 & 82 & 3 & \cmark & 0 & 19 \\
expenditures & Retail & 3 & 2 & 19 & 2 & \xmark & 0 & 2 \\
fnhk & Medicine & 3 & 2 & 21 & 2 & \xmark & 0 & 2 \\
gosales & Retail & 4 & 3 & 17 & 2 & \xmark & 0 & 3 \\
grants & Education & 12 & 11 & 30 & 4 & \xmark & 2 & 10 \\
lahman & Sport & 25 & 31 & 319 & 6 & \cmark & 6 & 28 \\
movielens & Entertain. & 7 & 6 & 14 & 4 & \xmark & 0 & 6 \\
restbase & Retail & 3 & 3 & 10 & 1 & \cmark & 2 & 2 \\
sakila & Retail & 16 & 23 & 75 & 7 & \cmark & 16 & 14 \\
sales & Retail & 4 & 3 & 12 & 2 & \xmark & 0 & 3 \\
sap & Retail & 4 & 3 & 35 & 2 & \xmark & 2 & 2 \\
seznam & Retail & 4 & 3 & 10 & 2 & \xmark & 0 & 3 \\
\midrule
\multicolumn{9}{c}{\textsc{RelBench} Databases} \\ 
\midrule
amazon & Retail & 3 & 2 & 10 & 2 & \xmark & 0 & 2 \\
avito & Retail & 8 & 11 & 23 & 2 & \cmark & 0 & 11 \\
f1 & Sport & 9 & 13 & 45 & 3 & \cmark & 0 & 13 \\
stack & Education & 7 & 12 & 33 & 3 & \cmark & 2 & 11 \\
trial & Medicine & 15 & 15 & 110 & 4 & \cmark & 4 & 13 \\
\bottomrule
\end{tabular}
    \caption{List of databases used in pretraining of the models.}
    \label{tab:db-info}
\end{table}

\subsection{Experimental Setup \& Benchmarks}
Our pretraining corpus consists of a diverse set of multi-tabular databases. We utilize the established RelBench~\cite{robinson_relbench_2024} datasets alongside a suite of 13 temporal databases from the CTU Relational~\cite{motl_ctu_2015} suite accessed through the Redelex framework~\cite{peleska_redelex_2025}. All used databases are listed in Table~\ref{tab:db-info} with their respective characteristics. For schema text (table/column names) and raw categorical tokens we use pooled GloVe vectors~\cite{pennington2014glove}. Structural neighborhood sampling is explicitly temporally constrained to prevent future information from entering message passing. In all settings we use batch size $128$.

\subsubsection{Pretraining}
The unified pretraining optimization is performed using the AdamW optimizer with a learning rate of $0.001$ and weight decay of $0.1$. To balance gradient scales across diverse pretraining tasks (Pillar 4), we employ a learning rate scheduler comprising a linear warmup phase starting at learning rate of $0.0001$ for the first 10\% of the maximum 50,000 training steps, followed by a Cosine Annealing decay.  Models are pretrained on a set of binary classification and regression tasks, where the target values of the regression tasks are normalized for the purposes of the pretraining regime. Tasks are randomly permuted to mitigate a bias of the fixed order while ensuring uniform representation.

To allow correct evaluation of foundational transferability, we employ a strict leave-one-database-out pretraining protocol. The models are pretrained on $N-1$ databases and subsequently evaluated on the held-out database. We are specifically using only the RelBench databases in place of the held-out database, meaning that the CTU Relational datasets are always used for the pretraining of the models.

\subsubsection{Evaluation}
To assess downstream performance, we train the models on tasks defined on the database held out during pretraining. Similarly to the pretraining phase, we use the AdamW optimizer with a learning rate of $0.001$ and a weight decay of $0.1$, training for only $5000$ steps to evaluate rapid adaptation. We report downstream performance using Receiver Operating Characteristic Area Under the Curve (ROC AUC) for binary classification and Mean Absolute Error (MAE), which are the standard evaluation metrics in RDL~\cite{robinson_relbench_2024,lachi_integrating_2025,wu_large_2025}.

\subsubsection{Models}
As the goal of this paper is not to train a single best model, but rather to demonstrate a modular approach to developing foundational models for RDL, we avoid extensive hyperparameter search. All models use fixed hyperparameters, with the only exception being the number of layers in the row encoder ($1$, $2$, or $4$). The Universal Row Encoder's transformer module utilizes $8$ attention heads with a dropout rate of $0.1$. All models employ a GraphSAGE~\cite{hamilton_inductive_2017} convolution backbone with $2$ message-passing layers and an inner embedding dimension of $512$. The task-specific baseline mirrors the prior model architecture~\cite{robinson_relbench_2024} often used as the RDL baseline~\cite{dwivedi_relational_2025,peleska_redelex_2025}, a heterogeneous GNN that employs a dedicated row encoder for each table, with learnable weights for each column based on its semantic type. Lastly, for all tasks, we employ a single fully connected layer to map the model's latent space to the prediction values, intentionally keeping the prediction head capacity small.

\subsection{Pretraining Universal Encoders}
\begin{figure}[t]
    \centering
    \includegraphics[width=\textwidth]{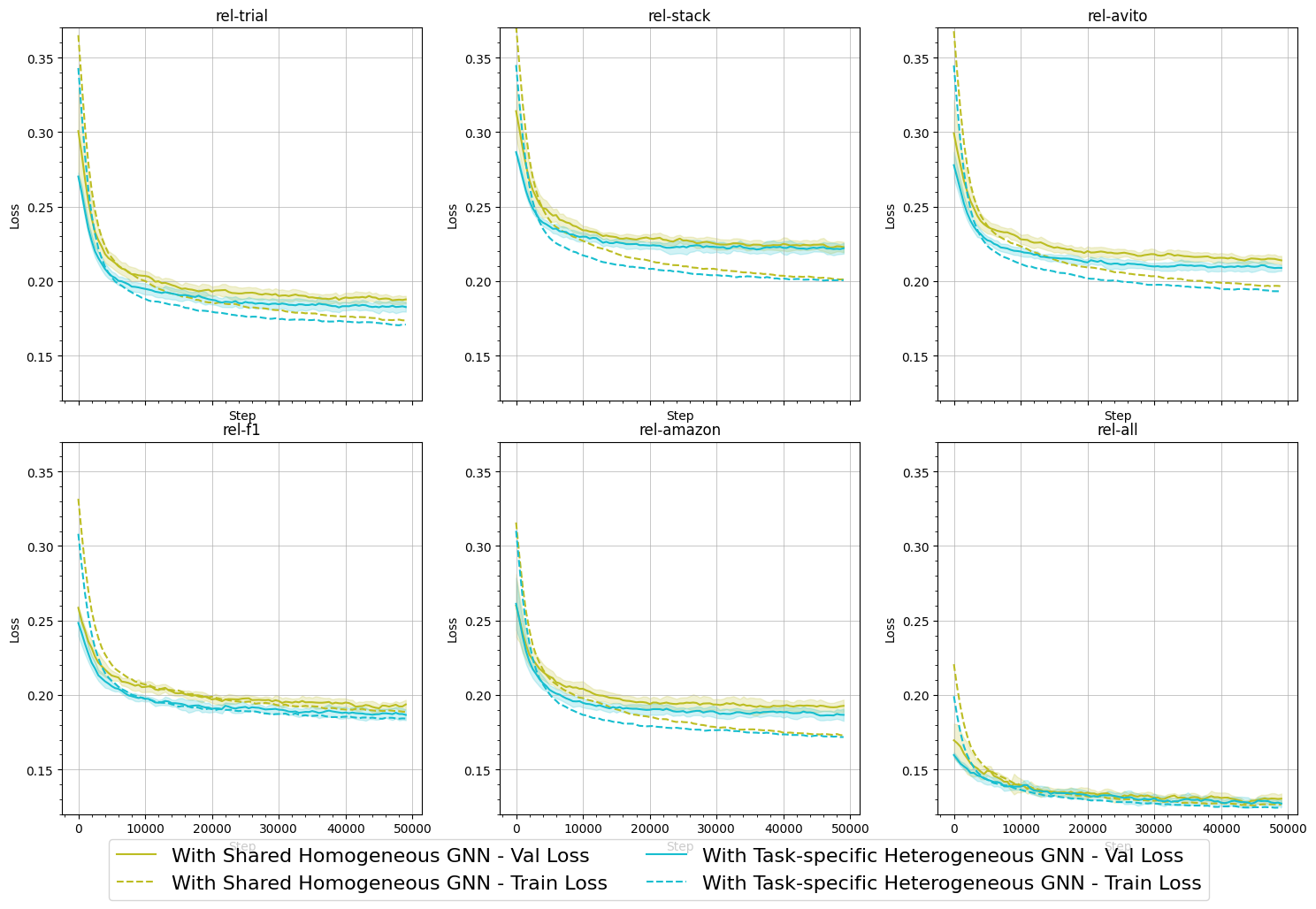}
    \caption{Pretraining loss trajectories across multiple hold-out scenarios. The plots compare the training and validation loss of the Universal Row Encoder coupled with a Shared Homogeneous GNN and a Task-specific Heterogeneous GNN. Both architectural variants demonstrate stable convergence across different hold-out scenarios.}
    \label{fig:encoder_pretraining}
\end{figure}
We compare two message-passing backbones attached to the Universal Row Encoder: a simplified Homogeneous GNN, with shared weights across node and edge types, and a task-specific Heterogeneous GNN with relation-dependent parameters.

Figure~\ref{fig:encoder_pretraining} shows that both variants converge stably across leave-one-database-out pretraining runs. The heterogeneous backbone achieves slightly lower loss trajectories, consistent with its higher relational capacity, while the homogeneous backbone remains competitive and notably stable across hold-out databases. We additionally include a \texttt{rel-all} setting (training only on CTU Relational while holding out all RelBench datasets) to contextualize loss scales under maximal domain shift.

\subsection{Transferability with Frozen Encoders}
\begin{figure}[t]
    \centering
    \includegraphics[width=\textwidth]{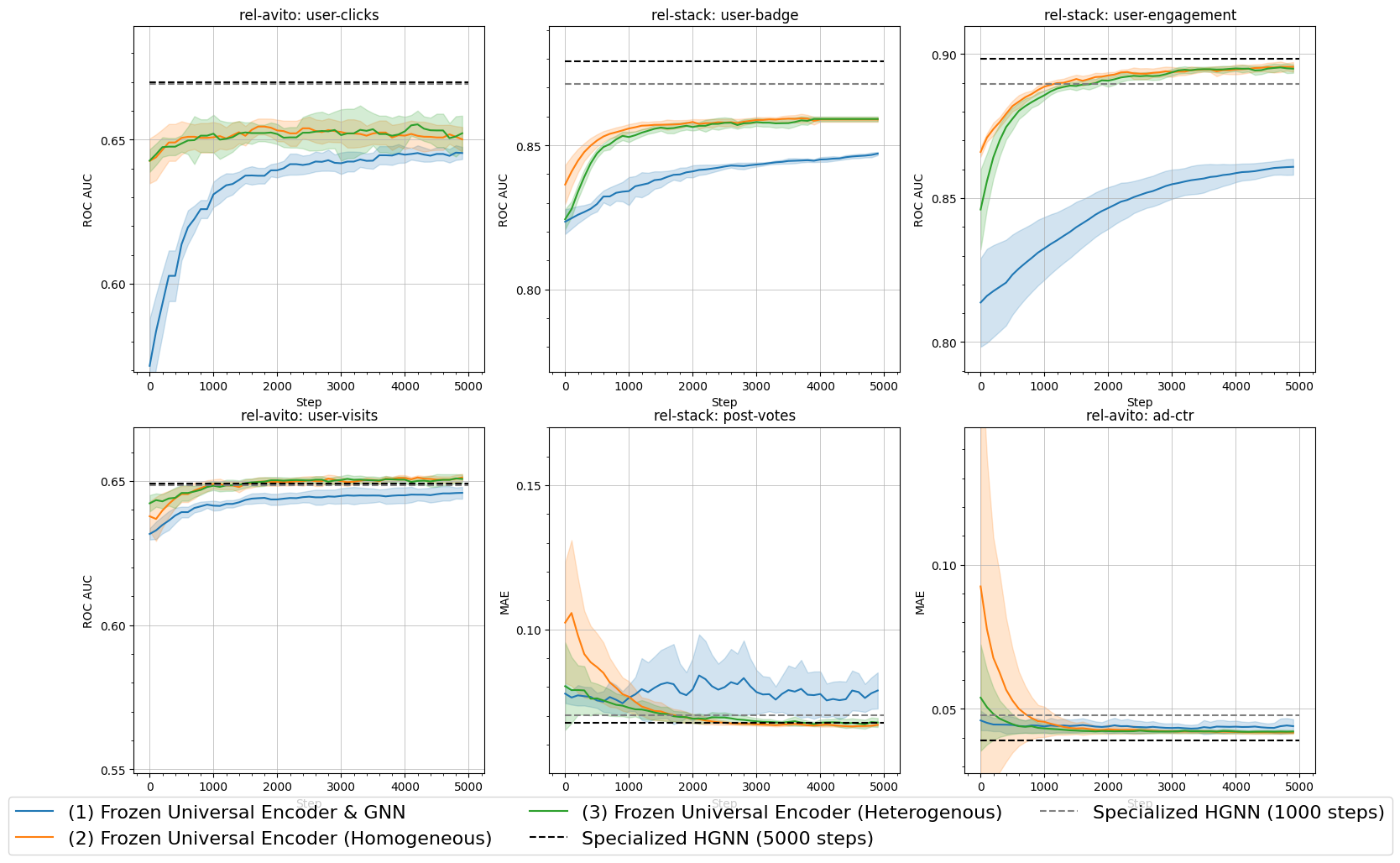}
    \caption{Downstream task evaluation of the Universal Row Encoder in different training regimes. The plots show the evaluation metrics (ROC AUC for classification, MAE for regression) on unseen databases across evaluation steps.}
    \label{fig:encoder_test}
\end{figure}
To evaluate transferability, we assess performance on databases held out from pretraining under three regimes: (1) \textit{head-only transfer}, where the encoder and homogeneous GNN are frozen and only a task head $d_{\psi_\ell}$ is trained; (2) \textit{frozen encoder}, where $f_{\theta}$ is fixed but the downstream GNN and head are trained; and (3) \textit{frozen encoder with task-specific pretraining}, where the encoder is fixed but was pretrained jointly with task-specific heterogeneous GNNs. We compare these to an end-to-end task-specific HGNN trained from scratch on the target database (after $1000$ and $5000$ steps).

Figure~\ref{fig:encoder_test} shows that head-only transfer is the most challenging regime, yet it can still provide non-trivial signal, particularly on some regression tasks. In contrast, regimes (2) and (3) exhibit consistent few-shot behavior on classification tasks and can match or surpass the task-specific baseline on selected tasks (e.g., \texttt{user-visits}, \texttt{post-votes}). Overall, these results suggest that a universal row encoder can reduce the burden of schema-specific input layers and improve rapid adaptation on unseen databases.

\subsection{Finetuning the Pretrained Models}
\begin{figure}[t]
    \centering
    \includegraphics[width=\textwidth]{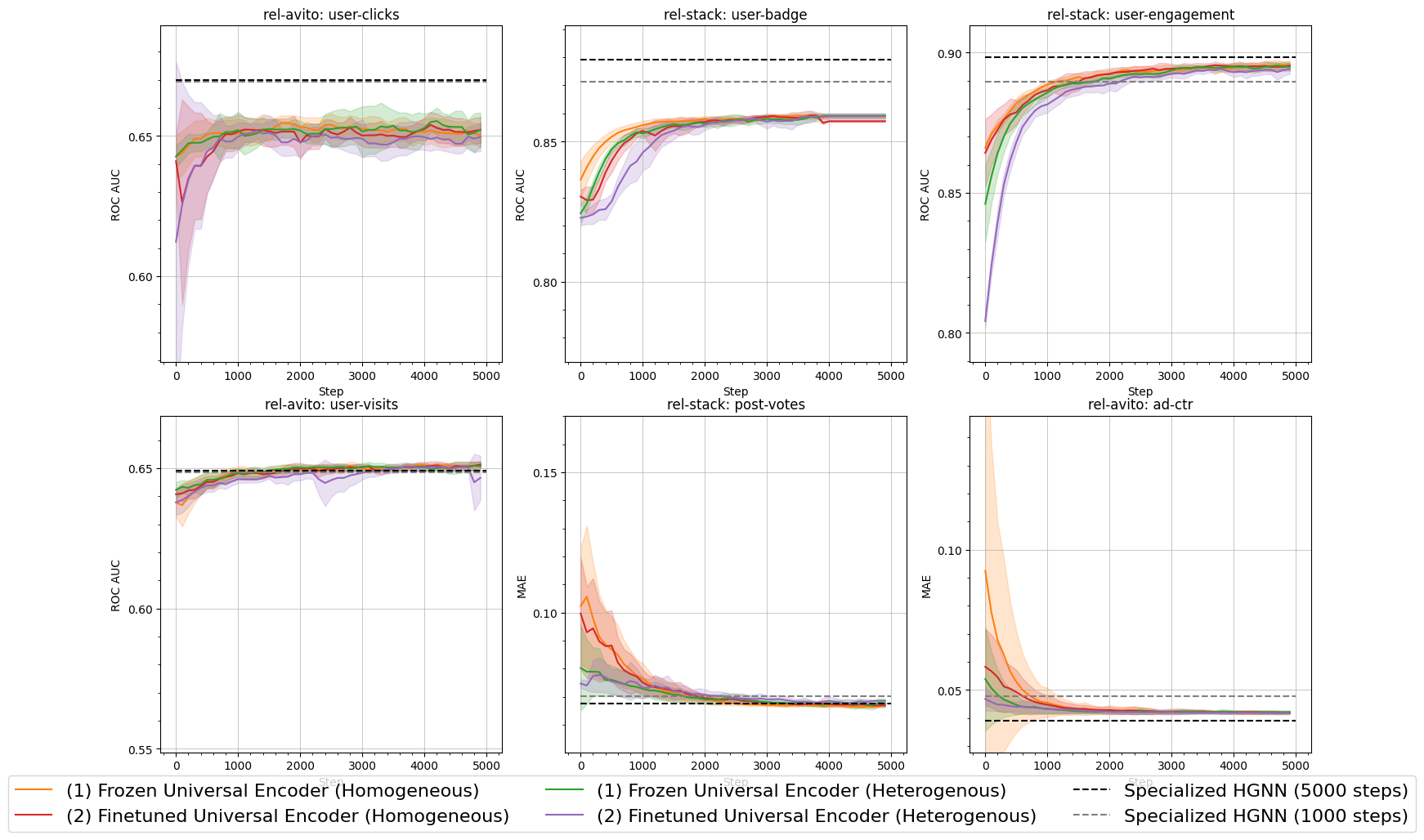}
    \caption{Comparison of frozen versus finetuned Universal Row Encoder regimes on downstream tasks. Finetuning the pretrained encoder alongside the GNN backbone yields minor to none additional performance relative to the completely frozen row encoder weights.}
    \label{fig:encoder_finetune}
\end{figure}

We compare keeping the pretrained row encoder frozen versus finetuning it jointly with the downstream GNN and task head. Figure~\ref{fig:encoder_finetune} shows only minor differences between the two regimes across the evaluated tasks and backbones. This suggests that most of the transferable benefit is already captured during pretraining, and that downstream training can often focus on adapting the structural module and task head. From a practical perspective, freezing $f_{\theta}$ simplifies deployment and reduces the computational cost of adapting the model to a new database.

\subsection{Memory-Performance Tradeoff}
\begin{table}[t]
    \centering
    \begin{tabular}{p{1.65cm}p{2.65cm}c p{1.5cm} p{1.5cm} p{1.5cm} p{1.5cm}}
\toprule
\multirow[c]{2}{*}{\textbf{Dataset}} & \multirow[c]{2}{*}{\textbf{Task}}  & \multirow[c]{2}{*}{~\textbf{Split}~}  & \multicolumn{2}{c}{\textbf{Specialized}} & \multicolumn{2}{c}{\textbf{Universal}} \\
 &  &  & MAE & Size MB & MAE & Size MB \\
\midrule
\multirow[t]{4}{*}{rel-amazon} & \multirow[t]{2}{*}{item-ltv} & val & 51.44 & 51.58 & 56.89 & 18.28 \\
 &  & test & 56.35 & 51.58 & 61.53 & 18.28 \\
\cline{2-7}
 & \multirow[t]{2}{*}{user-ltv} & val & 13.19 & 51.58 & 14.15 & 18.28 \\
 &  & test & 15.55 & 51.58 & 16.79 & 18.28 \\
\cline{1-7}
\multirow[t]{2}{*}{rel-avito} & \multirow[t]{2}{*}{ad-ctr} & val & 0.03 & 121.83 & 0.04 & 34.06 \\
 &  & test & 0.04 & 121.83 & 0.04 & 34.06 \\
\cline{1-7}
\multirow[t]{2}{*}{rel-f1} & \multirow[t]{2}{*}{driver-position} & val & 3.20 & 179.03 & 3.29 & 23.54 \\
 &  & test & 3.62 & 179.03 & 4.05 & 23.54 \\
\cline{1-7}
\multirow[t]{2}{*}{rel-stack} & \multirow[t]{2}{*}{post-votes} & val & 0.06 & 135.19 & 0.06 & 18.28 \\
 &  & test & 0.06 & 135.19 & 0.07 & 18.28 \\
\cline{1-7}
\multirow[t]{4}{*}{rel-trial} & \multirow[t]{2}{*}{site-success} & val & 0.36 & 368.81 & 0.45 & 34.06 \\
 &  & test & 0.39 & 368.81 & 0.47 & 34.06 \\
\cline{2-7}
 & \multirow[t]{2}{*}{study-adverse} & val & 48.27 & 368.81 & 54.37 & 34.06 \\
 &  & test & 47.04 & 368.81 & 55.79 & 34.06 \\
\cline{1-7}
\bottomrule
\end{tabular}

    \caption{Comparison of the of the Specialized model---schema-specific row encoder with HGNN---vs. the Universal model---Universal Row Encoder with Homogeneous GNN and frozen weigh---in terms of the best results on the regression downstream tasks and memory footprint (model size in MB). The Universal approach yields significant memory efficiency gains (up to 10$\times$ smaller footprint) at the cost of a mild drop in predictive accuracy.}
    \label{tab:model_size}
\end{table}

We evaluate the memory efficiency of our universal approach compared to specialized task-specific baselines. Table~\ref{tab:model_size} reports the predictive performance (MAE) alongside the model footprint (in MB) across various regression tasks. As expected, replacing a highly parameterized, schema-specific architecture with a single frozen universal encoder involves a practical trade-off: the results demonstrate a significant memory efficiency gain at the cost of a mild drop in MAE accuracy. 

Across the reported regression tasks, the Specialized model consistently\linebreak achieves lower error rates. For instance, on the \texttt{rel-trial} dataset, the Universal model exhibits a roughly $18\%$ relative increase in MAE compared to the Specialized baseline. However, this accuracy cost comes with a notable reduction in model size. On \texttt{rel-trial}, the Universal model requires only $34.06$ MB compared to $368.81$ MB for the Specialized model---a nearly $10\times$ reduction. Such reductions are important when deploying models across many databases, where per-schema encoders quickly become computationally intractable.


The reduction follows from parameter sharing: specialized baselines scale roughly with one row encoder per table, $|\Theta_{\mathrm{spec}}|\approx\sum_{k=1}^{K}|\theta^{(k)}_{\mathrm{row}}|+|\phi|$, whereas the universal approach reuses a single encoder, $|\Theta_{\mathrm{uni}}|\approx|\theta|+|\phi|$.

\section{Related Work}

The RDL paradigm was formalized~\cite{zahradnik_deep_2023,fey_position_2024} to connect deep learning and relational databases. The field is supported by benchmarks~\cite{wang_4dbinfer_2024,gu_relbench_2026,robinson_relbench_2024} as well as practical frameworks for building temporal relational graphs~\cite{peleska_redelex_2025}.

A large body of work explores architectures for learning on relational graphs, ranging from schema-specific relational GNNs~\cite{chen_relgnn_2025} to approaches that reuse strong tabular learners~\cite{lachi_boosting_2025,peleska_tabular_2025}. More recently, graph transformers and LLM-inspired models have been adapted to relational settings~\cite{dwivedi_relational_2025,wu_large_2025,lachi_integrating_2025}.

In parallel, the community is slowly moving toward \textit{foundational} RDL models trained across many databases~\cite{wehrstein_towards_2025,wang_griffin_2025,dwivedi_relational_2025,peleska_task-agnostic_2025}. These efforts often rely on monolithic trans\-former-based designs that jointly model cell-level semantics and relational propagation. Our work is complementary: we keep the relational message-passing module flexible, but focus on a universal, distribution-aware row encoder as a reusable backend. Synthetic relational data generation~\cite{hudovernik_deep_2025,kothapalli_plurel_2026} and further formalization of predictive RDL tasks~\cite{kocijan_predictive_2026} are likely to be important enablers for scaling such pretraining regimes.

\section{Conclusion}
In this work, we identified generalization bottlenecks in Relational Deep Learning (RDL) and distilled four pillars for relational foundation models: semantic granularity, structural topology, temporal causality, and unified optimization. To address these, we proposed a modular RDL architecture that decouples row-level semantic encoding from graph-level message passing.
We then introduced a new baseline implementation of the approach, with a Universal Row Encoder at its core. The encoder combines type-specific value embeddings with schema text and column-level distribution statistics to produce a fixed-size row representation that can be reused across different databases. Empirically, we showed that pretrained row encoding can accelerate adaptation on held-out databases and significantly reduce model sizes relative to schema-specific baselines, at the cost of but a mild drop in predictive accuracy.

\subsubsection{Limitations.} 
Our four-pillar framework establishes a conceptual roadmap for RDL foundation models; however, our current technical implementation prima\-rily addresses the first pillar (semantic granularity) and the schema-agnostic requirement of the second pillar. The remaining aspects of structural topology and temporal causality are currently handled via data representation mecha\-nisms---such as explicit primary-foreign key graphs and temporally constrained\linebreak sampling---rather than being intrinsically embedded within the model's architecture. Furthermore, regarding the fourth pillar (unified optimization), our approach was limited to target value normalization across just two task types: binary classification and regression. We did not explore loss balancing for broader task categories, such as multi-label classification or link prediction, nor did we implement a unified prediction head. Nevertheless, formalizing these pillars provides a necessary, unified vocabulary to guide the resolution of these bottlenecks in future research beyond Universal Row Encoder.

\subsubsection{Future Directions.}
While our baseline demonstrates the viability of a modular approach, several immediate avenues for refinement remain. First, future work should formally ablate the contextualization components in Equation~\ref{eq:contextualization}---namely the global statistics, schema names, and missingness indicators---alongside the intra-row pooling strategies (Equation~\ref{eq:pooling-strategies}) to rigorously isolate their individual contributions. Second, while precomputed summary statistics serve as an effective heuristic, exploring more principled mechanisms such as learned quantile features or percentile positions will be vital for capturing complex, non-parametric distributions. Similarly, replacing lightweight GloVe vectors with modern sentence encoders (e.g., MiniLM~\cite{wang2020minilm}, E5~\cite{wang2024multilingual}) promises straightforward performance gains.

Looking beyond the encoder, our framework opens broader research directions. To fully realize the desiderata defined by the four pillars, subsequent work must address the challenge of scaling model capacity~\cite{dwivedi_relational_2025}, potentially by leveraging synthetic relational data~\cite{kothapalli_plurel_2026,hollmann_accurate_2025} to satisfy the requirements for substantially larger, multi-domain training corpora~\cite{vogel_wikidbs_2024}.


\begin{credits}
\subsubsection*{Acknowledgments.} This work has received funding from the Czech Science Foundation grant No. 26-22501S. Computational infrastructure was provided by the OP VVV funded project CZ.02.1.01/0.0/0.0/16\_019/0000765 ``Research Center for Informatics''.

\subsubsection*{Ethical Considerations.} As foundational Relational Deep Learning (RDL) models scale by training on massive, multi-domain databases, they inherently risk capturing and propagating sensitive information. Real-world relational databases frequently contain Personally Identifiable Information (PII) and protected attributes across high-stakes domains like healthcare, finance, and criminal justice. The deployment of universal row encoders introduces the risk of unintended data memorization. Furthermore, models explicitly relying on global column statistics may inadvertently encode and amplify structural biases present in historical data distributions. Future research scaling these models must prioritize privacy-preserving pretraining objectives and develop robust auditing tools to ensure foundational RDL models are deployed equitably and securely.

\subsubsection*{Generative AI Statement.} Generative AI tools were used to assist in preparing this paper. They were employed responsibly to uphold the integrity of the submission, specifically restricted to enhancing the readability of the text.

\end{credits}

\bibliographystyle{splncs04}
\bibliography{bibliography}
\end{document}